\RequirePackage{amsmath}

\documentclass[runningheads]{llncs}
\usepackage[T1]{fontenc}
\usepackage{graphicx}
\usepackage{amsmath}
\usepackage{amssymb}
\usepackage{mathtools}
\usepackage{amsthm}
\usepackage{microtype}
\usepackage{graphicx}
\usepackage{subfigure}
\usepackage{booktabs} 
\usepackage{xcolor}
\usepackage{tikz-cd} 

\providecommand{\R}{\mathbb{R}}

\providecommand{\Id}{\mathrm{Id}}
\providecommand{\dd}{\mathrm{d}}

\providecommand{\tr}{\mathrm{tr}}

\providecommand{\End}{\mathrm{End}}

\providecommand{\IG}{\mathrm{IG}}
\providecommand{\RIG}{\mathrm{RIG}}

\newcommand\numberthis{\addtocounter{equation}{1}\tag{\theequation}}

\begin{document}
\title{Riemannian Integrated Gradients: A Geometric View of Explainable AI}
%
%
\author{Federico Costanza\inst{1}\and
Lachlan Simpson\inst{2}}
\authorrunning{F. Costanza and L. Simpson.}
%
\institute{Center for Theoretical Physics, Polish Academy of Sciences, Poland.
\and
School of Electrical and Mechanical Engineering, The University of Adelaide, Australia.}
\maketitle              

\begin{abstract}
We introduce Riemannian Integrated Gradients (RIG); an extension of Integrated Gradients (IG) to Riemannian manifolds. We demonstrate that RIG restricts to IG when the Riemannian manifold is Euclidean space. We show that feature attribution can be phrased as an eigenvalue problem where attributions correspond to eigenvalues of a symmetric endomorphism.
\end{abstract}

\section{Introduction}
The predictive power of deep learning comes with the trade-off of explainability \cite{zednik2019solving}. Explainability methods address this problem by providing an attribution of the input features to the prediction of a neural network. There is a long-standing hypothesis that data lies on a low-dimensional Riemannian manifold embedded in $\R^n$ \cite{fefferman2013testing,whiteley2024statisticalexplorationmanifoldhypothesis}.

Recent work has demonstrated that designing an explainability method which respects the geometry of the data manifold leads to more robust and intuitive explanations \cite{brodt,simpson-ig-alg,pmlr-v196-xenopoulos22a,zaher2024manifoldintegratedgradientsriemannian}. Analysis of the gradient of a neural network reveals the salient features of a prediction \cite{smilkov2017smoothgrad,pmlr-v70-sundararajan17a}. Explainability methods which utilise the gradient are aptly termed gradient explainability methods. Integrated Gradients \cite{pmlr-v70-sundararajan17a} is a popular gradient explainability method. Integrated gradients depends on a hyper-parameter known as the base-point which defines a path in the data-space. In \cite{simpson-tangential} the authors demonstrate that choosing the base-point of IG such that IG aligns with the tangent space at the point to explain provides user-friendly explanations. Zaher et al. \cite{zaher2024manifoldintegratedgradientsriemannian} demonstrate that if the path in IG is a geodesic within the embedded Riemannian manifold then the explanations are more robust to adversarial attack. 

A limitation of previous work is the assumption that the data manifold is embedded in $\R^n$ and has a specific geometric structure. Recent work has demonstrated that data may lie in non-Euclidean space such as the Poincaré half-plane $\mathbb{H}$ \cite{NEURIPS2018_dbab2adc}.

In this work, we define gradient explainability methods on a Riemannian manifold. Defining explainability in an abstract setting allows one to build an explainability method suited to the different geometries of the data.

The rest of the article is structured as follows: Section 2 defines gradient explainability methods in Euclidean space $\mathbb{R}^n$. Section 3 extends the definition of gradient explainability methods to Riemannian manifolds. We demonstrate that many of the axioms of IG break down in non-Euclidean spaces and must be adapted accordingly. Furthermore, Theorem 1 demonstrates that RIG restricts to IG in Euclidean space. Explainablity methods depend on a choice of basis. Under appropriate choice of basis, we demonstrate that RIG attributions correspond to the eigenvalues of a symmetric endomorphism; providing a rich geometric understanding of attributions.  

\section{Gradient Attribution Methods on Euclidean Space}
In this section, we define gradient-based attribution methods (GAM) in Euclidean space. We extend the usual definition of attribution methods to depend on an orthonormal basis. Furthermore, the axioms of baseline attribution methods defined in \cite{lundstrom2022rigorous} are generalised to this setting.

Explainability methods measure the extent each feature contributes to the prediction of a neural network. Explainability methods may be generalised as functions of the following form. 
\begin{definition}
\label{def:xai_def}
 Consider $\R^{n}$ equipped with an inner product $\langle \cdot, \cdot \rangle$. An explainability method is a map of the form
\begin{equation*}
    A : \R^{n} \times O(n) \times C^{1}(\R^{n}) \to \R^{n}, \quad A(x, U, F) = (A_{u_{1}}(x, F), \dots, A_{u_{n}}(x, F)), 
\end{equation*}  
where $u_{i}$ is the $i$-th column of $U$ and $A_{u_{i}} (x, F)$ denotes the attribution of $x$ to the prediction $F(x)$ in the direction $u_{i}$. 
\end{definition}

Generally, we will define attribution methods in terms of the attributions in the direction of a unit length vector. When $A_{u}(x, F)$ is a function of the directional derivative $\langle \nabla F (x),u \rangle$, we will say that $A$ is a \textit{gradient explainability method}. A \textit{base-line attribution method} (BAM) is a gradient explainability method that, in addition, is a function of a path $\gamma :[a,b] \to \R^n$. Given a base-point $x'$ and a point $x$ one constructs a path $\gamma :[a,b] \to \R^n$ with endpoints $x'$ and $x$. Explanations are made relative to a base-point $x'$. In \cite{simpson-ig-alg} the authors generalise BAMs to be coordinate-free:
\begin{equation}
    A^{\gamma}_{u}(x,F) = \int_{a}^{b} \langle \nabla F(\gamma(t)), u \rangle \langle \gamma'(t), u \rangle \; \dd t.
\end{equation}
In this article we focus on IG, a specific BAM where $\gamma$ is the straight line between $x'$ and $x$. IG is defined as:
\begin{equation}\label{def:IG_u}
\mathrm{IG}_{u}(x, x', F) : = \langle x - x', u \rangle  \int_{0}^{1} \langle \nabla F (x' + t(x - x')), u \rangle \; \dd t.
\end{equation}
In this article, we extend this definition to the case when $\R^{n}$ is replaced by a compact connected Riemannian manifold. 

GAMs and particularly, BAMs seek to satisfy several desirable axioms first introduced in \cite{pmlr-v70-sundararajan17a}. Below, the GAM axioms are generalised to attribution methods with respect to an orthonormal basis. We refer the reader to \cite{lundstrom2022rigorous} for an in-depth discussion of the following axioms:

\begin{enumerate}
    \item[\textbf{I}] Implementation invariance: If two neural networks are functionally equivalent, the attributions are the same.

    \item[\textbf{L}] Linearity: $A(x, aF + bG) = a A(x, F) + b A(x, G)$, for all $a, b \in \R$, $x\in \R^{n}$ and $F, G \in C^{1}(\R^{n})$.
    \item [\textbf{S}] Sensitivity: If $\langle u, \nabla F \rangle = 0$, then $A_{u}(x, F) = 0$ for all $x \in \R^{n}$.
    \item[\textbf{SI}] Symmetry invariance: For any pair $(i, j)$, let $s_{ij} : \R^{n} \to \R^{n}$ be the linear map such that $s_{ij}(u_{i}) = u_{j}$, $s_{ij}(u_{j}) = u_{i}$ and $s_{ij}(u_{k}) = u_{k}$, $k \neq i, j$. If $F(x) = F(s_{ij}(x))$ for all $x\in \R^{n}$, then 
    \begin{equation*}
        A_{u_{i}}(x, F) = A_{u_{j}}(s_{ij}(x), F).
    \end{equation*}
    \item[\textbf{C}] Completeness: For all $F \in C^{1}(\R^{n})$ and $x \in \R^{n}$, we have 
    \begin{equation*}
        \sum_{i = 1}^{n} A_{u_{i}}(x, F) = F(x) + \epsilon(F),
    \end{equation*}
    where $\epsilon(F) \in \R$ is an error term depending on $F$.
\end{enumerate}

\begin{remark}
    The error term introduced in Axiom \textbf{C} is $\epsilon(F) = -F(x')$ for IG.
\end{remark}

Sundararajan et al. \cite{pmlr-v70-sundararajan17a} claim that IG is the unique BAM which satisfies all of the aforementioned axioms. We will show in Theorem 2 that in fact all BAMs in Euclidean space satisfy these axioms.

\section{Gradient Explainability Methods on Riemannian Manifolds}
In this section, we provide a novel extension of attribution methods to Riemannian manifolds. First, we define attribution methods in Riemannian manifolds. Second, the axioms of Section 2 are adapted to Riemannian manifolds. Third, we introduce Riemannian Integrated Gradients (RIG), a novel generalisation of IG to a compact connected Riemannian manifold. We will use the notation introduced in Section~2 and, without loss of generality, will always consider neural networks in $C^{\infty}(M)$, the space of smooth functions from $M$ to $\R$.


In order to provide a generalisation of gradient explainability methods in Definition \ref{def:xai_def}, we will consider elements in $O(TM, g)$ as pairs $(p, U)$, where $p \in M$ and $U = (u_{1}, \dots, u_{n})$ is an orthonormal basis of $T_{p}M$.

\begin{definition}
    Let $(M, g)$ be a Riemannian manifold of dimension $n$. A gradient-based attribution method on $(M, g)$ is a map $A : O(TM, g) \times C^{\infty}(M) \to \R^{n}$
    \begin{equation}
        A(p, U, F) = (A_{u_{1}}(p, F), \dots, A_{u_{n}}(p, F)), 
    \end{equation}
    where $A_{u_{i}}(p, F)$ denotes the attribution in the direction of $u_{i}$, that is a function of $u_{i}(F)$.
\end{definition}

In non-Euclidean spaces, all but one GAM axiom can be naturally extended. The symmetry invariance axiom takes advantage of the vector space structure of $\R^{n}$ to ``swap directions'' that do not affect the neural network. Below we introduce an analogous axiom adapted to our geometric setting.
\begin{enumerate}
    \item[\textbf{II}] Isometry invariance: Let $s : M \to M$ be an isometry of $(M, g)$. Then
    \[
    A_{\dd s_{p} u}(s(p), F \circ s^{-1}) = A_{u}(p, F),
    \]
    for all $(p, u) \in TM$.
\end{enumerate}
Colloquially, this axiom states that transformations that preserve the Riemannian manifold structure (isometries), also preserve the attributions provided by the method. Particularly, if $(M, g)$ is Euclidean space and $s = s_{ij}$ as in Axiom \textbf{SI}, noting that $s_{ij}$ is a linear map such that $s_{ij}^{-1} = s_{ij}$, it is immediate that both axioms coincide when $F \circ s_{ij} = F$.

\subsection{Riemannian Integrated Gradients}

In the following, we will always assume that $(M, g)$ is a compact connected Riemannian manifold and therefore, by the Hopf-Rinow theorem, any two points in $M$ can be connected by a length-minimising geodesic. $F \in C^{\infty}(M)$ will denote a neural network and $o \in M$ a fixed base-point. For each point $p \in M$ to explain, $\gamma : [0, 1] \to M$ will always denote a smooth curve such that $\gamma(0) = p$ and $\gamma(1) = o$. $V(\gamma)$ will denote the vector space of vector fields along $\gamma$. Lastly, $P_{\gamma(t)} : T_{p}M \to T_{\gamma(t)}M$ will denote the parallel transport along $\gamma$, and for $u \in T_{p}M$, $P_{\gamma}u$ will denote the vector field along $\gamma$ with value $P_{\gamma(t)}u$ at $\gamma(t)$. 

Consider the bilinear map $\mathcal{A}_{F, \gamma} : V(\gamma) \times V(\gamma) \to \R$ given by
\begin{equation}\label{def:A_Fgamma}
	\mathcal{A}_{F, \gamma}(U, V) \coloneqq - \int_{\gamma} \dd F(U) g(V, \cdot ), \quad U, V \in V(\gamma).
\end{equation}
The above bilinear map naturally generalises BAMs to non-Euclidean geometries.  It is worth mentioning that in the literature, paths are usually taken from the base-point to the point to explain. We have chosen the opposite convention and corrected our definition of $\mathcal{A}_{F, \gamma}$ with a minus sign to account for this discrepancy. In Euclidean space, for any constant vector field $u$ in $\R^{n}$, it is immediate that $\mathcal{A}_{F, \gamma} (u, u) = A_{u}^{\gamma}(p, F)$ as defined in equation (1). This leads us to introduce another bilinear map in terms of $\mathcal{A}_{F, \gamma}$, defined point-wise as 
\begin{equation}\label{def:alpha_F}
    \alpha_{F, \gamma}(p)(u, v) : = \mathcal{A}_{F, \gamma}(P_{\gamma}u, P_{\gamma}v),
\end{equation}
where $\gamma$ is always a curve such that $\gamma(0) = p$. We shall refer to it as the \textit{path attribution form}.

\begin{remark}
    By construction, all BAMs in Euclidean space are defined by the path attribution form.
\end{remark}

\begin{definition}
    Let $(M, g, o)$ be a compact connected Riemannian manifold with fixed base-point $o \in M$. Riemannian Integrated Gradients with base-point $o$ is the gradient attribution method $\RIG : O(TM, g) \times C^{\infty}(M) \to \R^{n}$
    \begin{equation}
	   \RIG(p, U, F): = (\RIG_{u_{1}}(p, F), \dots, \RIG_{u_{n}}(p, F)), 
    \end{equation}
    with attribution in the direction of $u \in T_{p}M$ given by
    \begin{equation}\label{def:GIG_u}
            \RIG_{u}(p, F): = \alpha_{F, \gamma}(p)(u, u),
    \end{equation}
    where $\gamma : [0, 1] \to M$ is a length-minimising geodesic from $p$ to $o$.
\end{definition}

We have noted in Remark 1 that all BAMs in Euclidean space are defined in terms of the path attribution form, particularly IG. The choice of defining RIG in terms of parallel vector fields along geodesics was made to require only a tangent vector at the point to explain, rather than a vector field along a curve. 

\begin{theorem}
    $\mathrm{RIG}$ coincides with $\mathrm{IG}$ in Euclidean space.
\end{theorem}
\begin{proof}
    Let $(M, g)$ be Euclidean space and $o$ be our base-point. It is enough to prove that for a unit vector $u \in T_{p}M$, the equality $\RIG_{u}(p, F) = \IG_{u}(p, o, F)$ holds. Parallel transport is trivial in Euclidean space, namely $P_{\gamma(t)} = \Id$, and under the identification of the tangent space of $\R^{n}$ with $\R^{n}$ itself we get 
    \begin{equation*}
        \dd F(P_{\gamma(t)}u) = g (\nabla F (\gamma(t)), P_{\gamma(t)}u) = g (\nabla F (\gamma(t)), u).
    \end{equation*}
    Also, in Euclidean space geodesics are straight lines, for which $\gamma'(t) = -(o-p)$.
    
    Lastly, it follows from the definition of RIG that
    \begin{equation*}
        \RIG_{u}(p, F) = - \int_{0}^{1} \dd F(P_{\gamma(t)}u) g(u, \gamma'(t)) \dd t
        = \int_{0}^{1} g (\nabla F (\gamma(t)), u) g(u, p-o) \dd t.
    \end{equation*}
    The right-hand side of the above equation is exactly $\IG_{u}(p, o, F)$ as per equation~(\ref{def:IG_u}).
\end{proof}
In order to address the Riemannian base-line axioms for Riemannian Integrated Gradients, we proceed to investigate properties of the path attribution form. Below, Proposition 1 and 2 address Axioms \textbf{II} and \textbf{C}, respectively.

\begin{proposition}
    Let $s : (M, g) \to (M, g)$ be an isometry. Then
    \begin{equation}
        \alpha_{F \circ s^{-1}, s \circ \gamma} (s(p))(\dd s_{p} u, \dd s_{p} v) = \alpha_{F, s}(p)(u, v)
    \end{equation}
    for all $u, v \in T_{p}M$.
\end{proposition}
\begin{proof}
    We want to prove that the following coincides with $\alpha_{F, \gamma}(p)(u, v)$
    \begin{align*}
         \alpha_{F \circ s^{-1}, s \circ \gamma} &(s(p))(\dd s_{p} u, \dd s_{p} v) \\= &- \int_{0}^{1} \dd (F \circ s^{-1}) (P_{(s \circ \gamma)(t)} \dd s_{p}u) g(P_{(s\circ\gamma)(t)}\dd s_{p}v, (s \circ \gamma)'(t)) \; \dd t \numberthis.
    \end{align*}
    By assumption $s$ is an isometry, for which the following diagram commutes
    \begin{equation*}
        \begin{tikzcd}
            T_{p}M \arrow[rrr, "\dd s_{p}"] \arrow[dd, "P_{\gamma(t)}"] &&& T_{s(p)}M \arrow[dd, "P_{(s \circ \gamma)(t)}"] \\
            &&& \\
T_{\gamma(t)}M &&& T_{(s \circ \gamma)(t)}M \arrow[lll, "\dd s^{-1}_{(s \circ \gamma)(t)}"]
\end{tikzcd}
    \end{equation*}
    or in other words $ \dd s^{-1}_{(s \circ \gamma)(t)} \circ P_{(s \circ \gamma)(t)} \circ \dd s_{p} = P_{\gamma(t)}$. It follows from the above diagram that
    \begin{equation*}
        \dd (F \circ s^{-1}) (P_{(s \circ \gamma)(t)} \dd s_{p}u) = \dd F_{(s^{-1}\circ s \circ \gamma)(t)} \dd s^{-1}_{(s \circ \gamma)(t)} P_{(s \circ \gamma)(t)} \dd s_{p} u = \dd F(P_{\gamma(t)}u)
    \end{equation*}
    and 
    \begin{equation*}
        g(P_{(s\circ\gamma)(t)}\dd s_{p}v, (s \circ \gamma)'(t)) = g(\dd s_{\gamma (t)} P_{\gamma(t)}v, \dd s_{\gamma(t)} \gamma'(t)) 
        = g(P_{\gamma(t)}v, \gamma'(t)).  
    \end{equation*}
    Replacing the above two equations into equation (11) we get the desired result.
\end{proof}

\begin{proposition}
    Let $(M, g)$ be a connected Riemannian manifold and $\gamma : [0, 1] \to M$ be a smooth curve such that $\gamma(0) = p$ and $\gamma(1) = o$. Then
     \begin{equation}
         \tr \; \alpha_{F, \gamma}(p) = F(p) - F(o).
     \end{equation}
\end{proposition}
\begin{proof}
    Let $\lbrace u_{i} \rbrace_{i = 1}^{n}$ be an orthonormal basis of $T_{p}M$ and define $\eta_{i} : = g(u_{i}, \cdot)$ for $i = 1, \dots, n$. Since $g$ is parallel, we have $g(P_{\gamma(t)}u_{i}, \cdot) = P_{\gamma(t)} \eta_{i}$ and therefore
     \begin{equation}\label{eqn:trace}
        \tr \; \alpha_{F, \gamma}(p) = \sum\limits_{i = 1}^{n} \alpha_{F, \gamma}(p)(u_{i}, u_{i}) = - \int_{\gamma} \sum\limits_{i = 1}^{n} \dd F(P_{\gamma}u) P_{\gamma}\eta.
    \end{equation}
    The integrand in equation (\ref{eqn:trace}) is nothing but $\dd F$ expressed in a local frame along $\gamma$. Therefore, it follows from Stokes theorem that
    \begin{equation*}
        \tr \; \alpha_{F, \gamma}(p) = - \int_{\gamma} \dd F = - (F(o) - F(p)).
    \end{equation*}
\end{proof}

BAMs defined by the path attribution form satisfy Axioms \textbf{I}, \textbf{L} and \textbf{S} trivially. Proposition 1 guarantees us that they satisfy Axiom \textbf{II}. Regarding Axiom \textbf{C}, suppose that $A^{\gamma}_{u}(p, F) = \alpha_{F, \gamma}(p)(u, u)$. Choosing an orthonormal basis $\lbrace u_{i} \rbrace_{i = 1}^{n}$ of $T_{p}M$, by Proposition 2 we have that
    \begin{equation*}
        \sum\limits_{i=1}^{n} A^{\gamma}_{u_{i}}(p, F) = \sum\limits_{i=1}^{n} \alpha_{F, \gamma}(p)(u_{i}, u_{i}) = \tr \; \alpha_{F, \gamma}(p) = F(p) - F(o).
    \end{equation*}
    Consequently, BAMs satisfy Axiom \textbf{C} with error term $\epsilon(F) = - F(o)$. We have proved the following theorem.

\begin{theorem}
     Base-line attribution methods defined by the path attribution form satisfy the Riemannian base-line attribution axioms.   
\end{theorem}

We note that whilst the above results are for a manifold $M$, when working with data, we must assume the manifold hypothesis. Shao et al. \cite{riemannian_geom} provide methods to compute the embedded manifold, metric, and length minimising geodesic. Utilising the work of Shao et al. RIG can be directly applied to a data manifold.

\subsection{A Natural Choice of Basis for Riemannian Integrated Gradients}
        BAMs defined by the path attribution form rely on choices of orthonormal basis. Each basis provides a different explanation. We aim to provide a natural choice of basis for each tangent space at the point to explain.
    
        It was implicitly hidden in Proposition 2 that the attributions given by $\alpha_{F, \gamma}$ are related to its eigenvalues. Let us consider the symmetrisation of the path attribution form:
		\begin{equation*}
		    \dot{\alpha}_{F, \gamma}(p)(u, v) = \frac{1}{2} (\alpha_{F, \gamma}(p)(u, v) + \alpha_{F, \gamma}(p)(v, u)).
		\end{equation*}
        Certainly, $\dot{\alpha}_{F, \gamma}$ defines the same BAM as $\alpha_{F, \gamma}$, since $\dot{\alpha}_{F, \gamma}(p)(u, u) = \alpha_{F, \gamma}(p)(u, u)$. With the aid of the metric tensor, we will let $Q_{F, \gamma} (p)\in \End (T_{p}M)$ be the endomorphism of $T_{p}M$ associated to $\dot{\alpha}_{F, \gamma}$, defined implicitly by
        \begin{equation*}
            \dot{\alpha}_{F, \gamma}(p)(u, v) = g(Q_{F, \gamma}(p)u, v).
        \end{equation*}

       The endomorphism $Q_{F, \gamma}(p)$ is symmetric. Consequently, its eigenvalues are real and its eigenvectors define orthogonal basis of $T_{p}M$. Choosing an orthonormal basis $\lbrace u_{i} \rbrace_{i = 1}^{n}$ of eigenvectors of $Q_{F, \gamma}(p)$, we have that the attribution in the direction of $u_{i}$ is precisely the eigenvalue $\lambda_{i}$ associated to $u_{i}$, namely 
	\[
	\alpha_{F, \gamma}(p)(u_{i}, u_{i}) = \lambda_{i}.
	\]
        Below we provide a bound on attributions in terms of the eigenvalues of $Q_{F, \gamma}(p)$.
        
        \begin{proposition}
            Let $\lbrace u_{i} \rbrace_{i=1}^{n}$ be an orthonormal basis of eigenvectors of $Q_{F, \gamma}(p)$ such that $|\lambda_{1}| \leq \dots \leq |\lambda_{n}|$. Then
            \[
            |\alpha_{F, \gamma}(p)(u, u)| \leq |\lambda_{n}|, 
            \]
            for all $u \in T_{p}M$ of unit length.
        \end{proposition} 
        \begin{proof}
            It follows directly from the triangle inequality.
        \end{proof}






\section{Conclusion}
In this work, explainability methods were abstracted to Riemannian manifolds. The axioms of base-line attribution methods were extended to Riemannian manifolds. RIG was introduced as a novel extension of IG to a connected compact Riemannian manifold. We demonstrated that RIG obeys axioms analogous to IG in the Riemannian setting and RIG restricts to IG when $M = \R^n$. Lastly, we showed that under appropriate choice of basis, RIG attributions are eigenvalues of the path attribution form. In future work, we seek to experimentally validate RIG on datasets with different geometries.
%
%
%
\bibliographystyle{splncs04}
\bibliography{mybibliography}

\begin{thebibliography}{10}
\providecommand{\url}[1]{\texttt{#1}}
\providecommand{\urlprefix}{URL }
\providecommand{\doi}[1]{https://doi.org/#1}

\bibitem{brodt}
Bordt, S., Uddeshya, U., Akata, Z., von Luxburg, U.: {T}he {M}anifold {H}ypothesis for {G}radient-{B}ased {E}xplanations. In: 2023 IEEE/CVF Conference on Computer Vision and Pattern Recognition Workshops (CVPRW). pp. 3697--3702 (2023)

\bibitem{fefferman2013testing}
Fefferman, C., Mitter, S., Narayanan, H.: {T}esting the {M}anifold {H}ypothesis. Journal of the American Mathematical Society  \textbf{29},  983–1049 (2016)

\bibitem{NEURIPS2018_dbab2adc}
Ganea, O., Becigneul, G., Hofmann, T.: {H}yperbolic {N}eural {N}etworks. In: Bengio, S., Wallach, H., Larochelle, H., Grauman, K., Cesa-Bianchi, N., Garnett, R. (eds.) Advances in Neural Information Processing Systems (NIPS) (2018)

\bibitem{lundstrom2022rigorous}
Lundstrom, D., Huang, T., Razaviyayn, M.: {A} {R}igorous {S}tudy of {I}ntegrated {G}radients {M}ethod and {E}xtensions to {I}nternal {N}euron {A}ttributions. Proceedings of the 39th International Conference on Machine Learning  \textbf{162},  14485--14508 (2022)

\bibitem{riemannian_geom}
Shao, H., Kumar, A., Fletcher, P.T.: {T}he {R}iemannian {G}eometry of {D}eep {G}enerative {M}odels. In: 2018 IEEE/CVF Conference on Computer Vision and Pattern Recognition Workshops (CVPRW). pp. 428--4288 (2018)

\bibitem{simpson-ig-alg}
Simpson, L., Costanza, F., Millar, K., Cheng, A., Lim, C.C., Chew, H.G.: {A}lgebraic {A}dversarial {A}ttacks on {I}ntegrated {G}radients. International Conference on Machine Learning and Cybernetics (ICMLC)  (2024), ({T}o {A}ppear)

\bibitem{simpson-tangential}
Simpson, L., Costanza, F., Millar, K., Cheng, A., Lim, C.C., Chew, H.G.: {T}angentially {A}ligned {I}ntegrated {G}radients for {U}ser-{F}riendly {E}xplanations. 32nd Irish Conference on Artificial Intelligence and Cognitive Science, (AICS)  (2024), ({T}o {A}ppear)

\bibitem{smilkov2017smoothgrad}
Smilkov, D., Thorat, N., Kim, B., Viégas, F., Wattenberg, M.: {S}moothgrad: {R}emoving {N}oise by {A}dding {N}oise. arXiv preprint arXiv:1706.03825  (2017)

\bibitem{pmlr-v70-sundararajan17a}
Sundararajan, M., Taly, A., Yan, Q.: {A}xiomatic {A}ttribution for {D}eep {N}etworks. Proceedings of the 34th International Conference on Machine Learning (ICML)  \textbf{70},  3319--3328 (2017)

\bibitem{whiteley2024statisticalexplorationmanifoldhypothesis}
Whiteley, N., Gray, A., Rubin-Delanchy, P.: {S}tatistical {E}xploration of the {M}anifold {H}ypothesis. arXiv preprint arXiv:2208.11665  (2024)

\bibitem{pmlr-v196-xenopoulos22a}
Xenopoulos, P., Chan, G., Doraiswamy, H., Nonato, L.G., Barr, B., Silva, C.: {GALE}: {G}lobally {A}ssessing {L}ocal {E}xplanations. In: Proceedings of Topological, Algebraic, and Geometric Learning Workshops 2022. Proceedings of Machine Learning Research (PMLR), vol.~196, pp. 322--331 (2022)

\bibitem{zaher2024manifoldintegratedgradientsriemannian}
Zaher, E., Trzaskowski, M., Nguyen, Q., Roosta, F.: {M}anifold {I}ntegrated {G}radients: {R}iemannian {G}eometry for {F}eature {A}ttribution. arXiv preprint arXiv:2405.09800  (2024)

\bibitem{zednik2019solving}
Zednik, C.: {S}olving the {B}lack {B}ox {P}roblem: {A} {N}ormative {F}ramework for {E}xplainable {A}rtificial {I}ntelligence. Philosophy \& Technology  \textbf{34},  265--288 (2021)

\end{thebibliography}
%




\end{document}